\documentclass[runningheads]{llncs}
\usepackage{graphicx}
\usepackage{tikz}
\usepackage{comment}
\usepackage{amsmath,amssymb} % define this before the line numbering.
\usepackage{color}
\usepackage[accsupp]{axessibility}  % Improves PDF readability for those with disabilities.

\usepackage[ruled]{algorithm2e}
\usepackage{multirow}
\usepackage{orcidlink}
\usepackage[misc]{ifsym}

\begin{document}
% \linenumbers
\pagestyle{headings}
\mainmatter
\def\ECCVSubNumber{5914}  % Insert your submission number here

\title{BoundaryFace: A mining framework with noise label self-correction  for Face Recognition} % Replace with your title

\begin{comment}
\titlerunning{ECCV-22 submission ID \ECCVSubNumber} 
\authorrunning{ECCV-22 submission ID \ECCVSubNumber} 
\author{Anonymous ECCV submission}
\institute{Paper ID \ECCVSubNumber}
\end{comment}
%******************

%\begin{comment}
\titlerunning{BoundaryFace}
\author{Shijie Wu\orcidlink{0000-0002-7125-752X} \and
Xun Gong{\Letter}\orcidlink{0000-0002-1494-0955}} 
\authorrunning{Shijie Wu, Xun Gong}
% \institute{Southwest Jiaotong University\\
% \email{sxu\_wushijie@163.com}\\
% \and
% Southwest Jiaotong University\\
% \email{xgong@home.swjtu.edu.cn}}

\institute{School of Computing and Artificial Intelligence, Southwest Jiaotong University, Chengdu, Sichuan, China\\
\email{xgong@swjtu.edu.cn}}
%\email{sxu\_wushijie@163.com, xgong@home.swjtu.edu.cn}}
%\end{comment}
%******************
\maketitle

\begin{abstract}
Face recognition has made tremendous progress in recent years due to the advances in loss functions and the explosive growth in training sets size. A properly designed loss is seen as key to extract discriminative features for classification. Several margin-based losses have been proposed as alternatives of softmax loss in face recognition. However, two issues remain to consider: 1) They overlook the importance of hard sample mining for discriminative learning. 2) Label noise ubiquitously exists in large-scale datasets, which can seriously damage the model's performance.  In this paper, starting from the perspective of decision boundary, we propose a novel mining framework that focuses on the relationship between a sample’s ground truth class center and its nearest negative class center. Specifically, a closed-set noise label self-correction module is put forward, making this framework work well on datasets containing a lot of label noise. The proposed method consistently outperforms SOTA methods in various face recognition benchmarks. Training code has been released at \url{https://github.com/SWJTU-3DVision/BoundaryFace}.
% \dots
\keywords{Face Recognition, Noise Label, Hard Sample Mining, Decision Boundary}
\end{abstract}

\section{Introduction}
Face recognition is one of the most widely studied topics in the computer vision community. Large-scale datasets, network architectures, and loss functions have fueled the success of Deep Convolutional Neural Networks (DCNNs) on face recognition. Particularly, with an aim to extract discriminative features, the latest works have proposed some intuitively reasonable loss functions.

For face recognition, the current existing losses can be divided into two approaches: one deems the face recognition task to be a general classification problem, and networks are therefore trained using softmax \cite{liu2017sphereface,wang2017normface,wang2018cosface,wang2018additive,deng2019arcface,cao2020domain,guo2020learning,ranjan2017l2,yuan2017feature}; the other approaches the problem using metric learning and directly learns an embedding, such as \cite{sun2015deep,schroff2015facenet,sohn2016improved}. Since metric learning loss usually suffers from sample batch combination explosion and semi-hard sample mining, the second problem needs to be addressed by more sophisticated sampling strategies. Loss functions have therefore attracted increased attention.

It has been pointed out that the classical classification loss function ($i.e$., Softmax loss) cannot obtain discriminative features. Based on current testing protocols, the probe commonly has no overlap with the training images, so it is particularly crucial to extract features with high discriminative ability. To this end, Center loss \cite{wen2016discriminative} and NormFace \cite{wang2017normface} have been successively proposed to obtain discriminative features. Wen $et\ al.$ \cite{wen2016discriminative}  developed a center loss that learns each subject's center. To ensure the training process is consistent with testing, Wang $et\ al.$ \cite{wang2017normface} made the features extracted by the network and the weight vectors of the last fully connected layer lay on the unit hypersphere. Recently, some margin-based softmax loss functions \cite{liu2017sphereface,wang2018cosface,wang2018additive,deng2019arcface,liu2016large} have also been proposed to enhance intra-class compactness while enlarging inter-class discrepancy, resulting in more discriminative features.

\begin{figure}[t]
    \centering
    \includegraphics[scale=0.35]{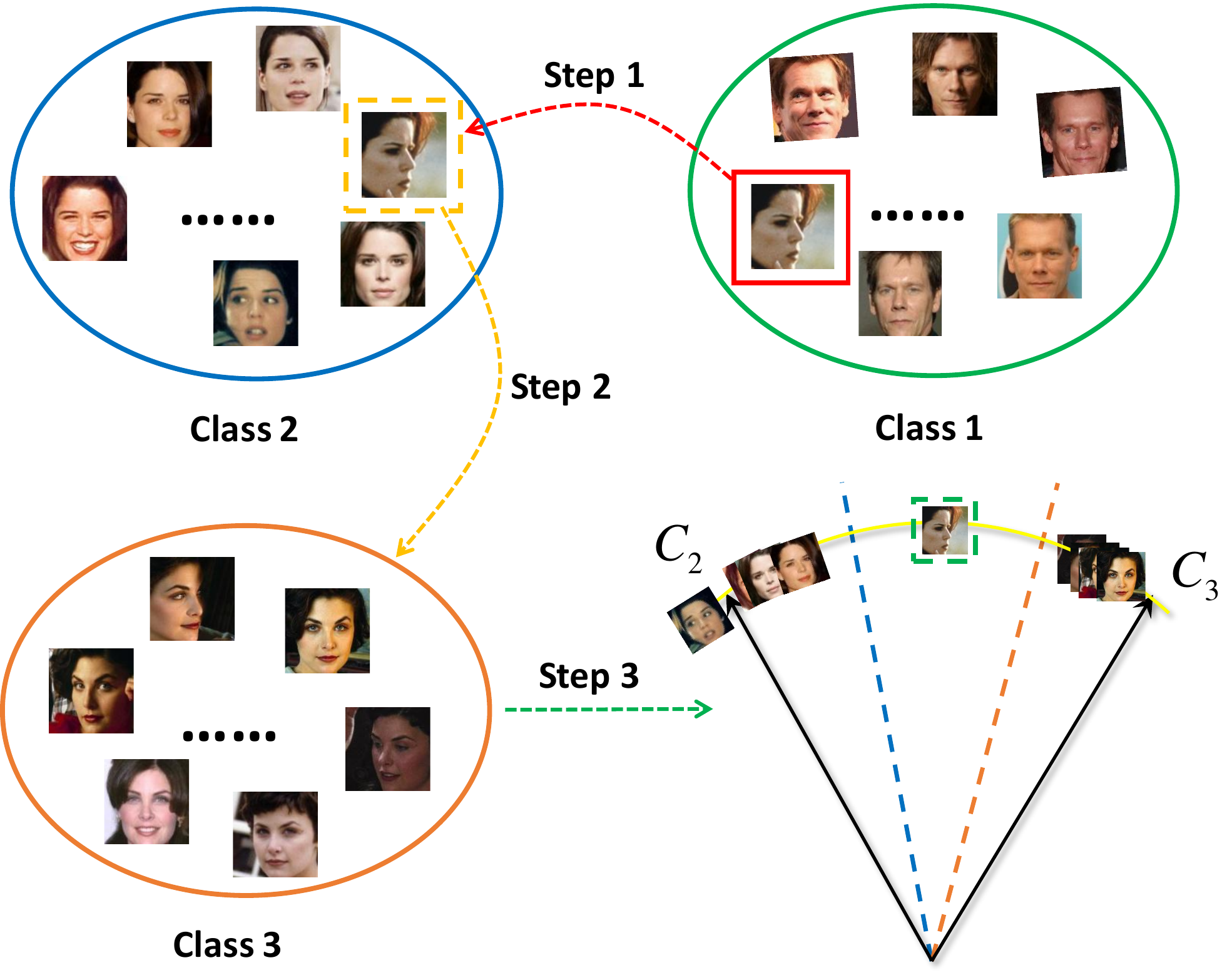}
    \caption{The motivation of BoundaryFace. Step 1 denotes closed-set noise label self-correction. Step 2 denotes nearest negative class match. Step 3 denotes hard sample handle. For a noisy hard sample, we first correct its label, then match the nearest negative class based on the correct label, and finally emphasize it using the decision boundary consisting of this sample's ground truth class center and the nearest negative class center.}
    \label{fig:motivation}
\end{figure}
The above approaches have achieved relatively satisfactory results. However, there are two very significant issues that must still be addressed: 1) Previous research has ignored the importance of hard sample mining for discriminative learning. As illustrated in \cite{huang2020improving,chen2020angular}, hard sample mining is a crucial step in improving performance. Therefore, some mining-based softmax losses have emerged. Very recently, MV-Arc-Softmax \cite{wang2020mis}, and CurricularFace \cite{huang2020curricularface} were proposed. They were inspired by integrating both margin and mining into one framework. However, both consider the relationship between the sample ground truth class and all negative classes, which may complicate the optimization of the decision boundary. 2) Both margin-based softmax loss and mining-based softmax loss ignore the influence of label noise. Noise in face recognition datasets is composed of two types: \textit{closed-set noise}, in which some samples are falsely given the labels of other identities within the same dataset, and \textit{open-set noise}, in which a subset of samples that do not belong to any of the classes, are mistakenly assigned  one of their labels, or contain some non-faces. Wang $et\ al.$ \cite{wang2018devil} noted that noise, especially closed-set noise, can seriously impact model's performance. Unfortunately, removing noise is expensive and, in many cases, impracticable. Intuitively, the mining-based softmax loss functions can negatively impact the model if the training set is noisy. That is, mining-based softmax is likely to perform less well than baseline methods on datasets with severe noise problems. Designing a loss function that can perform hard sample mining and tolerate noise simultaneously is still an open problem.

In this paper, starting from the perspective of decision boundary, we propose a novel mining framework with tolerating closed-set noise. Fig. \ref{fig:motivation} illustrates our motivation using a noisy hard sample processing. Specifically, based on the premise of closed-set noise label correction, the framework directly emphasizes hard sample features that are between the ground truth class center and the nearest negative class center. We find out that if a sample is a closed-set noise, there is a high probability that the sample is distributed within the nearest negative class's decision boundary, and the nearest negative class is likely to be the ground truth class of the noisy sample. Based on this finding, we propose a module that automatically discovers closed-set noise during training and dynamically corrects its labels. Based on this module, the mining framework can work well on large-scale datasets under the impact of severe noise. To sum up, the contributions of this work are:

\begin{itemize}
    \item We propose a novel mining framework with noise label self-correction, named BoundaryFace, to explicitly perform hard sample mining as a guidance of the discriminative feature learning.
    \item The closed-set noise module can be used in any of the existing margin-based softmax losses with negligible computational overhead. To the best of our knowledge, this is the first solution for closed-set noise from the perspective of the decision boundary.
    \item We have conducted extensive experiments on popular benchmarks, which have verified the superiority of our BoundaryFace over the baseline softmax and the mining-based softmax losses.
\end{itemize}
%------------------------------------------------------------------------
\section{Related Work}
%-------------------------------------------------------------------------
\subsection{Margin-based softmax}
Most recently, researchers have mainly focused on designing loss functions in the field of face recognition. Since basic softmax loss cannot guarantee  facial features that are sufficiently discriminative, some margin-based softmax losses \cite{liu2016large,liu2017sphereface,wang2018cosface,wang2018additive,deng2019arcface,zhang2019adacos}, aiming at enhancing intra-class compactness while enlarging inter-class discrepancy, have been proposed. Liu $et\ al.$ \cite{liu2016large} brought in multiplicative margin to face recognition in order to produce discriminative feature. Liu $et\ al.$ \cite{liu2017sphereface} introduced an angular margin (A-Softmax) between ground truth class and other classes to encourage larger inter-class discrepancy. Since multiplicative margin could encounter optimization problems, Wang $et\ al.$ \cite{wang2018cosface} proposed an additive margin to stabilize optimization procedure. Deng $et\ al.$ \cite{deng2019arcface} changed the form of the additive margin, which generated a loss with clear geometric significance. Zhang $et\ al.$ \cite{zhang2019adacos} studied on the effect of two crucial hyper-parameters of traditional margin-based softmax losses, and proposed the AdaCos, by analyzing how they modulated the predicted clasification probability. Even these margin-based softmax losses have achieved relatively good performance, none of them takes into account the effects of hard sample mining and label noise. 

%-------------------------------------------------------------------------
\subsection{Mining-based softmax}
There are two well-known hard sample mining methods, \textit{i.e.}, Focal loss \cite{lin2017focal}, Online Hard Sample Mining (OHEM) \cite{shrivastava2016training}. Wang $et\ al.$ \cite{wang2020mis}  has shown that naive combining them to current popular face recognition methods has limited improvement. Some recent work, MV-Arc-Softmax \cite{wang2020mis}, and CurricularFace \cite{huang2020curricularface} are inspired by integrating both margin and mining into one framework. MV-Arc-Softmax explicitly defines mis-classified samples as hard samples and adaptively strengthens them by increasing the weights of corresponding negative cosine similarities, eventually producing a larger feature margin between the ground truth class and the corresponding negative target class. CurricularFace applies curriculum learning to face recognition, focusing on easy samples in the early stage and hard samples in the later stage. However, on the one hand, both take the relationship between the sample ground truth class and all negative classes into consideration, which may complicate the optimization of the decision boundary; on the other hand, label noise poses some adverse effect on mining. It is well known that the success of face recognition nowadays benefits from large-scale training data. Noise is inevitably in these million-scale datasets. Unfortunately, Building a “clean enough” face dataset, however, is both costly and difficult. Both MV-Arc-Softmax and CurricularFace assume that the dataset is clean (\textit{i.e.}, almost noiseless), but this assumption is not true in many cases. Intuitively, the more noise the dataset contains, the worse performance of the mining-based softmax loss will be. Unlike open-set noise, closed-set noise can be part of the clean data as soon as we correct their labels. Overall, our method differs from the currently popular mining-based softmax in that our method can conduct hard sample mining along with the closed-set noise well being handled, while the current methods cannot do so.

%-------------------------------------------------------------------------
% add NPT-Loss
\subsection{Metric learning loss}
Triplet loss\cite{schroff2015facenet} is a classical metric learning algorithm. Even though the problem of combinatorial explosion has led many researchers to turn their attention to the adaptation of traditional softmax, there are still some researchers who explore the optimization of metric loss. Introducing the idea of proxy to metric learning is the mainstream choice at present. The proxy-triplet\cite{wang2017normface,MovshovitzAttias2017NoFD} replaces the positive and negative samples in the standard triplet loss with positive and negative proxies. Very recently, NPT-Loss\cite{Khalid2021NPTLossAM} has been proposed as a further modification of the proxy-triplet. NPT-Loss replaces the negative proxies in the proxy-triplet with nearest-neighbour negative proxy and the final form does not contain any hyper-parameters. NPT-Loss also has the effect of implicitly hard-negative mining. BoundaryFace's motivation is also partially inspired by NPT-Loss. The key differences between our BoundaryFace with NPT-Loss are: 1) NPT-Loss is essentially a metric learning loss, while BoundaryFace is essentially a softmax-based loss. 2) NPT-Loss still suffers from label noise. Intuitively, although metric learning algorithm can achieve the intra-class compactness and inter-class discrepancy more directly, the noise problem may be more prominent. 3) Even with implicit mining effects, NPT-Loss does not explicitly semanticize hard samples. In terms of the final formula, NPT-Loss treats all samples fairly. In contrast, starting from the perspective of decision boundary, based on the premise of closed-set noise label correction, BoundaryFace directly emphasizes hard sample features that are located in the margin region. 
% 三元组损失是一个经典的度量学习算法。即使组合爆炸的问题导致很多研究者把注意力转向对传统softmax的改造，
% 但是仍有一些研究者探究对度量损失的优化。将代理的思想引入到度量学习算法是目前主流的选择。代理三元组将
% 标准三元组损失中的正负样本改变为正负代理。非常最近，NPT-Loss被提出作为对代理三元组的进一步改造。NPT-Loss将代理三元组中的负代理改为最近邻负代理并且最终的形式不含有任何超参数。NPT-Loss也有隐式地困难样本挖掘的效果。
% BoundaryFace也受到NPT-Loss的部分鼓舞。在BoundaryFace和NPT-Loss之间的关键区别是：1)NPT-Loss本质上是一个度量学习损失，而BoundaryFace本质上是一个softmax-based损失。 2)NPT-Loss虽然有隐式地挖掘效果，但仍然会遭受噪声样本的
% 困扰。直觉上，虽然度量学习的算法可以更直接地达到类内聚，类间开的效果，但是噪声问题可能会更加突出。 
% 3) 即使有隐式地挖掘效果，NPT-Loss并没有明确地语义化困难样本。从最终公式来看，NPT-Loss公平地对待所有的样本。
%-------------------------------------------------------------------------
\section{The Proposed Approach}
%-------------------------------------------------------------------------
\subsection{Preliminary Knowledge}
\noindent{\bf Margin-based softmax.} The original softmax loss formula is as follows:
\begin{gather}
    L =  - \log \frac{{{e^{{W_{{y_i}}}{x_i} + {b_{{y_i}}}}}}}{{\sum\limits_{j = 1}^n {{e^{{W_j}{x_i} + {b_j}}}} }}
    \label{eq:1}
\end{gather}
where $ {x_i} $ denotes the feature of the $i$-th sample belonging to  ${y_i}$ class in the min-batch, ${W_j}$ denotes the $j$-th column of the weight matrix $W$ of the last fully connected layer, and ${b_j}$ and $n$ denote the bias term and the number of identities, respectively.

To make the training process of the face recognition consistent with the testing, Wang $et\ al.$ \cite{wang2017normface} let the weight vector $W_j$ and the sample features $ {x_i} $ lie on a hypersphere by $l_2$ normalization. And to make the networks converge better, the sample features are re-scaled to $s$. Thus, Eq. \ref{eq:1} can be modified as follows:
\begin{gather}
    L =  - \log \frac{{{e^{s(\cos {\theta _{{y_i}}})}}}}{{\sum\limits_{j = 1}^n {{e^{s(\cos {\theta _j})}}} }}
    \label{eq:2}
\end{gather}

With the above modification, ${W_j}$ has a clear geometric meaning which is the class center of $j$-th class and we can even consider it as a feature of the central sample of $j$-th class. ${\theta _{{{\rm{y}}_i}}}$ can be seen as the angle between the sample and its class center, in particular, it is also the geodesic distance between the sample and its class center from the unit hypersphere perspective. However, as mentioned before, the original softmax does not yield discriminative features, and the aforementioned corrections (\textit{i.e.}, Eq. \ref{eq:2}) to softmax do not fundamentally fix this problem, which has been addressed by some variants of softmax based on margin. They can be formulated in a uniform way:
\begin{gather}
    L =  - \log \frac{{{e^{sf(\cos {\theta _{{y_i}}})}}}}{{{e^{sf(\cos {\theta _{{y_i}}})}} + \sum\limits_{j = 1,j \ne {y_i}}^n {{e^{s(\cos {\theta _j})}}} }}
\end{gather}
E.g, in baseline softmax ($e.g.$, ArcFace), $f(\cos {\theta _{{y_i}}}) = \cos ({\theta _{{y_i}}} + m)$. As can be seen, the currently popular margin-based softmax losses all achieve intra-class compactness and inter-class discrepancy by squeezing the distance between a sample and its ground truth class center.

~\\
{\bf Mining-based softmax.} Hard sample mining is to get the network to extra focus valuable, hard-to-learn samples. There are two main categories in the existing mining methods that are suitable for face recognition: 1) focusing on samples with large loss values from the perspective of loss. 2) focusing on samples mis-classified by the network from the relationship between sample ground truth class and negative classes. They can be formed by a unified formula as below:
\begin{gather}
    L =  - I(p({x_i}))\log \frac{{{e^{sf(\cos {\theta _{{y_i}}})}}}}{{{e^{sf(\cos {\theta _{{y_i}}})}} + \sum\limits_{j = 1,j \ne {y_i}}^n {{e^{sg(t,\cos {\theta _j})}}} }}
\end{gather}
where $p({x_i}) = \frac{{{e^{sf(\cos {\theta _{{y_i}}})}}}}{{{e^{sf(\cos {\theta _{{y_i}}})}} + \sum\limits_{j = 1,j \ne {y_i}}^n {{e^{sg(t,\cos {\theta _j})}}} }}$ is the predicted ground truth probability and $I(p({x_i}))$ is an indicator function. For type 1, such as Focal loss, $I(p({x_i})) = {(1 - p({x_i}))^\lambda }$, $f(\cos {\theta _{{y_i}}}) = \cos {\theta _{{y_i}}}$ and $g(t,\cos {\theta _j}) = \cos {\theta _j}$, $\lambda$ is a modulating factor. For type 2, MV-Arc-Softmax and CurricularFace handle hard samples with varying $g(t,\cos {\theta _j})$. Let $N = f(\cos {\theta _{{y_i}}}) - \cos {\theta _j}$, thus, MV-Arc-Softmax is described as:
\begin{gather}
      g(t,\cos {\theta _j}) = {\begin{cases} {\cos {\theta _j}}, & {N \ge {\rm{0}}} \\
{\cos {\theta _j} + t}, & {N < 0}\end{cases}}
\end{gather}
and CurricularFace formula is defined as follows:
\begin{gather}
      g(t,\cos {\theta _j}) = {\begin{cases} {\cos {\theta _j}}, & {N \ge {\rm{0}}} \\
\cos {\theta _j}(t + \cos {\theta _j}), & {N < 0}\end{cases}}
\end{gather}
From the above formula, we can see that if a sample is a easy sample, then its negative cosine similarity will not change. Otherwise, its negative cosine similarity will be amplified. Specially, in MV-Arc-Softmax, ${\cos {\theta _j} + t} > \cos {\theta _j}$ always holds true since $t$ is a fixed hyper parameter and is always greater than 0. That is, the model always focuses on hard samples. In contrast, $t$ is calculated based on the Exponential Moving Average (EMA) in CurricularFace, which is gradually changing along with iterations. Moreover, the ${\cos {\theta _j}}$ can reflect the difficulty of the samples, and these two changes allow the network to learn easy samples in the early stage and hard samples in the later stage.
%------------------------------------------------------------------------
\begin{figure}[t]
    \centering
    \includegraphics[scale=0.35]{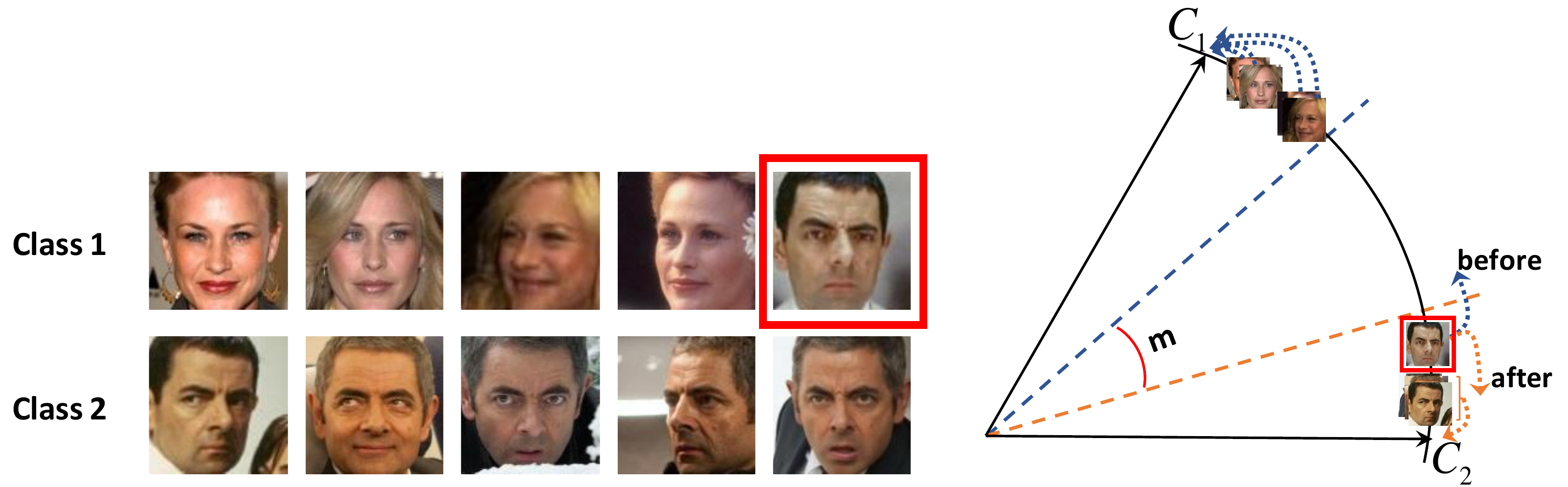}
    \caption{{\bf Left:} Each row represents one person. The red box includes a closed-set noise sample. {\bf Right:} The distribution of the samples in the left figure is shown from the perspective of the decision boundary. The dashed arrows represent the optimized direction of the samples.}
    \label{fig:my_label}
\end{figure}
\subsection{Label self-correction}
In this section, we discuss the mining framework's noise tolerance module. Unlike open-set noise, a closed-set noise sample is transformed into a clean sample if its label can be corrected appropriately. The existing mining-based softmax losses, as their prerequisite, assume that the training set is a clean dataset. Suppose the labels of most closed-set noise in a real noisy dataset are corrected; in that case, poor results from hard sample mining methods on noisy datasets can be adequately mitigated. More specifically, we find that when trained moderately, networks have an essential ability for classification; and that closed-set noise is likely to be distributed within the nearest negative class's decision boundary. Additionally, the negative class has a high probability of being the ground truth class of this sample. 
% \begin{figure}[t]
%     \centering
%     \includegraphics[scale=0.25]{boudnaryF1_v2.pdf}
%     \caption{{\bf Left:} Each row represents one person. The red box includes a closed-set noise sample. {\bf Right:} The distribution of the samples in the left figure is shown from the perspective of the decision boundary. The dashed arrows represent the optimized direction of the samples.}
%     \label{fig:my_label}
% \end{figure}
As shown in Fig. \ref{fig:my_label}, the red box includes a closed-set noise sample, which is labeled as class 1, but the ground truth label is class 2. The closed-set noise will be distributed within the decision boundary of class 2. At the same time, we dynamically change that sample's label so that the sample is optimized in the correct direction. That is, before the label of this closed-set noise is corrected, it is optimized in the direction of $C_1$ (\textit{i.e.}, the "before" arrow); after correction, it is optimized in the direction of $C_2$ (\textit{i.e.}, the "after" arrow). The label self-correction formula, named BoundaryF1, is defined as follows:
\begin{gather}
L =  - \frac{1}{N}\sum\limits_{i = 1}^N {\log \frac{{{e^{s\cos ({\theta _{y_i}} + m)}}}}{{{e^{s\cos ({\theta _{y_i}} + m)}} + \sum\limits_{j = 1,j \ne y_i}^n {{e^{s\cos {\theta _j}}}} }}} 
\label{eq:7}
\end{gather}
where $\ if \ \max \{ \cos ({\theta _k} + m) \ {\rm{ }}for{\rm{ }} \ all{\rm{ }} \ k \ne {y_i}\}  - \cos {\theta _{{y_i}}} > 0:{y_i} = k$. It means that, before each computing of Eq. \ref{eq:7}, we decide whether to correct the label based on whether the sample is distributed within the decision boundary of the nearest negative class. Noted that, as a demonstration, apply this method to ArcFace. It can be applied to other margin-based losses could also be used.
\begin{figure}[t]
    \centering
    \includegraphics[scale=0.25]{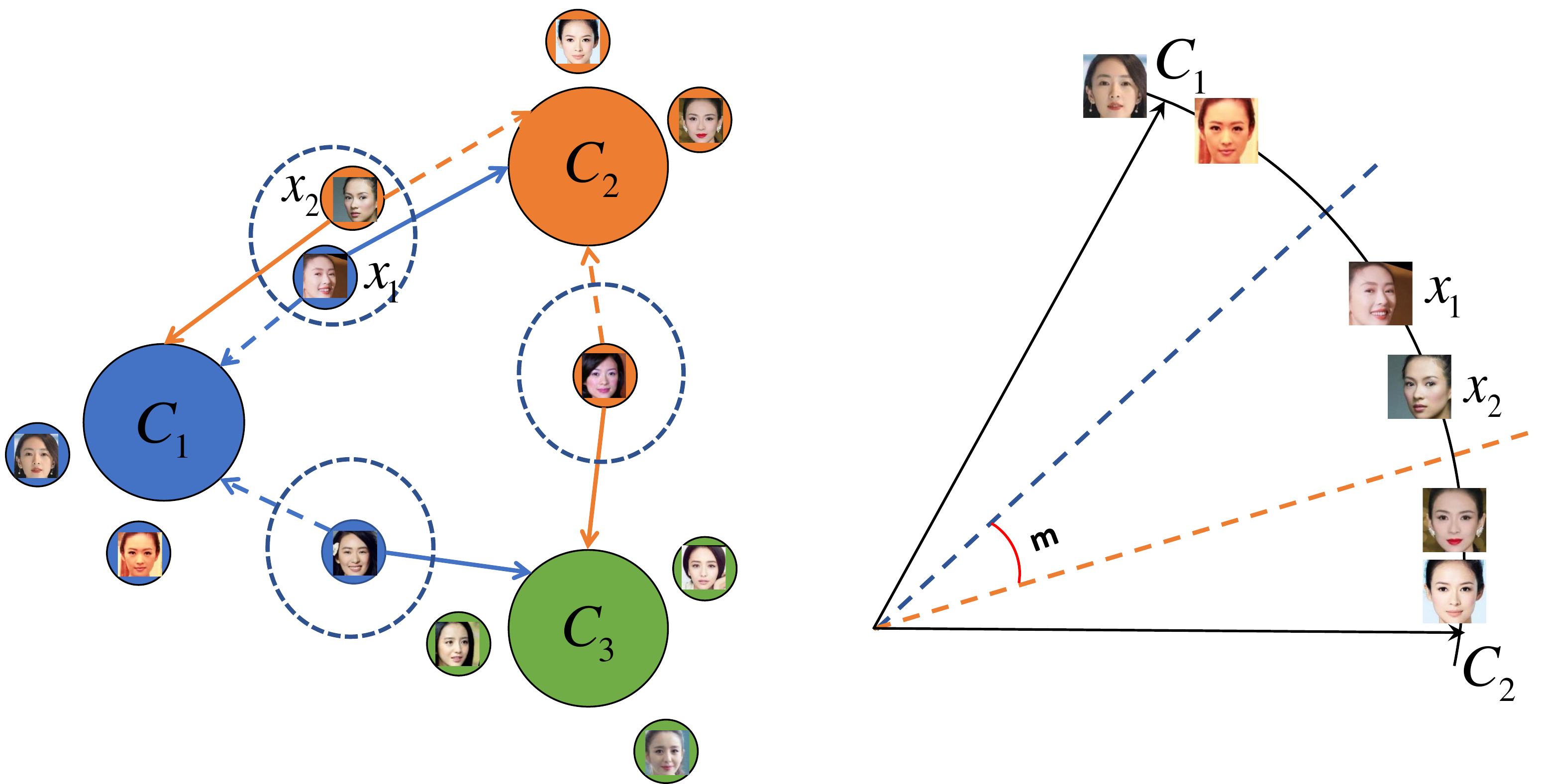}
    \caption{{\bf Left:} Blue, orange, and green represent three individuals. The samples in the ellipse are the hard samples. The solid arrows represent distance maximization, and the dashed arrows represent distance minimization.  {\bf Right:} The sample distribution of two persons in the left from the perspective of decision boundary.}
    \label{fig:my_label2}
\end{figure}
\subsection{BoundaryFace}
\begin{figure*}[pt]
    \centering
    \includegraphics[scale=0.35]{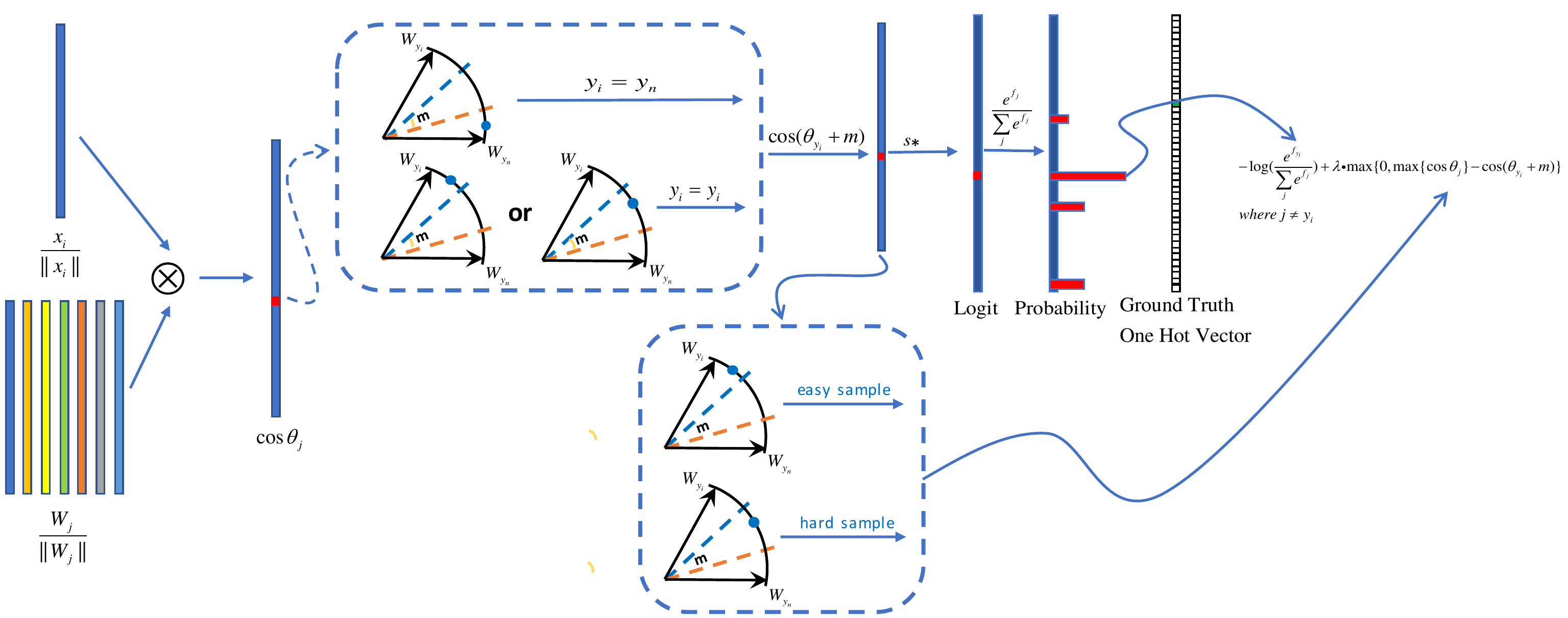}
    \caption{Overview of BoundaryFace. The part included in the upper dashed box represents the closed-set noise processing. 
    %When the sample is distributed within the decision boundary of the nearest negative class (\textit{i.e.}$, blue sample is distributed within the decision boundary of the orange class), we dynamically change their labels to $y_n$. 
    The part included in the lower dashed box represents the judgment of whether it is a hard sample or not. 
    %If the sample is distributed within the margin region, we consider it as a hard sample.
    }
    \label{fig:my_label3}
\end{figure*}

Unlike the mining-based softmax's semantic for assigning hard samples, we only consider samples located in the margin region between the ground truth class and the nearest negative class. In other words, as each sample in the high-dimensional feature space has a nearest negative class center, if the sample feature is in the margin region between its ground truth class center and the nearest negative class center, then we label it as a hard sample. As shown in Fig. \ref{fig:my_label2} (left), the nearest negative class for each class's sample may be different ($e.g.$, the nearest negative classes of two samples belonging to the $C_1$ class are $C_2$ and $C_3$, respectively). In Fig. \ref{fig:my_label2}, the right image presents the two classes of the left subfigure from the perspective of the decision boundary. 
%Noted that the center for the nearest negative class of sample $x_1$ is $C_2$ does not mean that its ground truth class center $C_1$ is the nearest to $C_2$. 
Since samples $x_1$ and $x_2$ are in the margin region between their ground truth class and the nearest negative class, we treat them as hard samples. An additional regularization term $f$ is added to allow the network to strengthen them directly. Additionally, to ensure its effectiveness on noisy datasets, we embed the closed-set noise label correction module into the mining framework. As shown in Fig. \ref{fig:my_label3}, after the network has obtained  discriminative power, for each forward propagation; and based on the normalization of feature $x_i$ and the weight matrix $W$, we obtain the cosine similarity $\cos {\theta _j}$ of sample feature $x_i$ to each class center $W_j$. 
% We find $\cos {\theta _{{y_i}}}$ of feature $x_i$ to the ground truth class (Corresponding to the its label) center ${W_{{y_i}}}$ and apply an angular margin penalty $m$ to it. After that, we get $\cos ({\theta _{{y_i}}} + m)$. 
Next, we calculate the position of feature $x_i$ at the decision boundary based on $\cos {\theta _j}$. Assuming that the nearest negative class of the sample is $y_n$. 
If it is distributed within the nearest negative class's decision boundary, we dynamically correct its label $y_i$ to $y_n$; otherwise proceed to the next step. We then calculate $\cos ({\theta _{{y_i}}} + m)$. After that, we simultaneously calculate two lines: one is the traditional pipeline; and the other primarily determines whether the sample is hard or not. These two lines contribute to each of the final loss function's two parts. Since our idea is based on the perspective of the decision boundary, we named our approach  BoundaryFace. The final loss function is defined as follows:
\begin{gather}
    L =  - \frac{1}{N}\sum\limits_{i = 1}^N {(\log \frac{{{e^{sT(\cos {\theta _{{y_i}}})}}}}{{{e^{^{sT(\cos ({\theta _{{y_i}}}))}}} + \sum\limits_{j = 1,j \ne {y_i}}^n {{e^{s\cos {\theta _j}}}} }}}  - \lambda f) \\
    T(\cos {\theta _{{y_i}}}) = \cos ({\theta _{{y_i}}} + m),\notag \\
   f = \max \{ 0,\max \{ \cos {\theta _j}{\rm{ |  }} \ for\ {\rm{ all  }}\ j \ne {y_i}\}  - T(\cos {\theta _{{y_i}}})\}\notag
\label{eq:8}
\end{gather}
where $\ if \ \max \{ \cos ({\theta _k} + m) \ {\rm{ }}for{\rm{ }} \ all{\rm{ }} \ k \ne {y_i}\}  - \cos {\theta _{{y_i}}} > 0:{y_i} = k$. $\lambda$ is a balance factor. As with BoundaryF1, before each computing of final loss, we decide whether to correct the label based on whether the sample is distributed within the decision boundary of the nearest negative class.
% boundaryF2 figure
% boundaryFace figure

~\\
{\bf Optimization} In this part, we show that out BoundaryFace is trainable and can be easily optimized by the classical stochastic gradient descent (SGD). Assuming $x_i$ denotes the deep feature of $i$-th sample which belongs to the $y_i$ class, $L_1 =  - \log (\frac{{{e^{{f_{{y_i}}}}}}}{{\sum\limits_k {{e^{{f_k}}}} }})$, $L_2 = \lambda \max \{ 0,\max \{ \cos {\theta _k}{\rm{ |  }} \ for \ {\rm{ all }} \ k \ne {y_i}\}  - \cos ({\theta _{{y_i}}} + m)\}$, the input of the $L_1$ is the logit $f_k$, where $k$ denotes the $k$-th class. 

In the forward propagation, when $k = {y_i}$, ${f_k} = s\cos ({\theta _{{y_i}}} + m)$, when $k \ne {y_i}$, ${f_k} = s\cos ({\theta _k})$. Regardless of the relationship of $k$ and $y_i$, there are two cases for $L_2$, if $x_i$ is a easy sample, $L_2 = 0$. Otherwise, it will be constituted as $L_2 = \lambda (\max \{ \cos {\theta _k}{\rm{ |  }} \ for \ {\rm{ all }} \ k \ne {y_i}\}  - \cos ({\theta _{{y_i}}} + m))$. In the backward propagation process, the gradients w.r.t. $x_i$ and $w_k$ can be computed as follows:
\begin{figure}[h]
    \centering
    \includegraphics[scale=0.45]{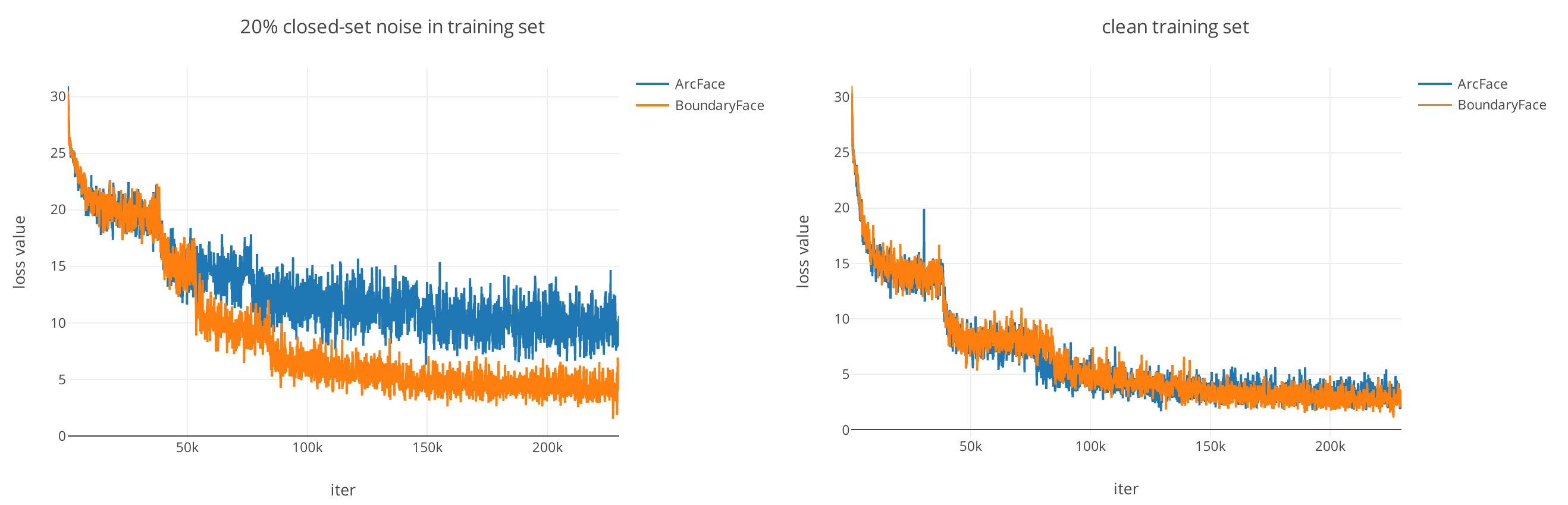}
    \caption{Convergence process of BoundaryFace.}
    \label{fig:my_label4}
\end{figure}
\\$when$ $k = {y_i}$
\begin{gather}
      \frac{{\partial L}}{{\partial {x_i}}} = {\begin{cases} \frac{{\partial L_1}}{{\partial {f_{{y_i}}}}}(s\frac{{\sin ({\theta _{{y_i}}} + m)}}{{\sin {\theta _{{y_i}}}}})\frac{{{w_{{y_i}}}}}{{||{w_{{y_i}}}||}}X, & {easy} \\
({\frac{{\partial L_1}}{{\partial {f_{{y_i}}}}}s + \frac{{\partial L_2}}{{\partial \cos ({\theta _{{y_i}}} + m)}})\frac{{\sin ({\theta _{{y_i}}} + m)}}{{\sin {\theta _{{y_i}}}}}\frac{{{w_{{y_i}}}}}{{||{w_{{y_i}}}||}}X}, & {hard}\end{cases}}
%\notag
\label{eq:9}
\end{gather}
\begin{gather}
      \frac{{\partial L}}{{\partial {w_k}}} = {\begin{cases} \frac{{\partial L_1}}{{\partial {f_{{y_i}}}}}(s\frac{{\sin ({\theta _{{y_i}}} + m)}}{{\sin {\theta _{{y_i}}}}})\frac{{{x_i}}}{{||{x_i}||}}W, & {easy} \\
({\frac{{\partial L_1}}{{\partial {f_{{y_i}}}}}s + \frac{{\partial L_2}}{{\partial \cos ({\theta _{{y_i}}} + m)}})\frac{{\sin ({\theta _{{y_i}}} + m)}}{{\sin {\theta _{{y_i}}}}}\frac{{{x_i}}}{{||{x_i}||}}W}, & {hard}\end{cases}}
%\notag
\label{eq:10}
\end{gather}
\\$when$ $k \ne {y_i}$
\begin{gather}
      \frac{{\partial L}}{{\partial {x_i}}} = {\begin{cases} \frac{{\partial L_1}}{{\partial {f_k}}}s\frac{{{w_k}}}{{||{w_k}||}}X, & {easy} \\
({\frac{{\partial L_1}}{{\partial {f_k}}}s + \frac{{\partial L_2}}{{\partial \cos {\theta _k}}})\frac{{{w_k}}}{{||{w_k}||}}X}, & {hard}\end{cases}}
%\notag
\label{eq:11}
\end{gather}
\begin{gather}
      \frac{{\partial L}}{{\partial {w_k}}} = {\begin{cases} \frac{{\partial L_1}}{{\partial {f_k}}}s\frac{{{x_i}}}{{||{x_i}||}}W, & {easy} \\
({\frac{{\partial L_1}}{{\partial {f_k}}}s + \frac{{\partial L_2}}{{\partial \cos {\theta _k}}})\frac{{{x_i}}}{{||{x_i}||}}W}, & {hard}\end{cases}}
\label{eq:12}
\end{gather}
where, both $X$ and $W$ are symmetric matrices.

Further, in Fig. \ref{fig:my_label4}, we give the loss curves of baseline and BoundaryFace on the clean dataset and the dataset containing 20\% closed-set noise, respectively. It can be seen that our method converges faster than baseline. The training procedure is summarized in Algorithm \ref{al:1}.
% 算法 伪代码
\begin{algorithm}[t]
  \caption{BoundaryFace}
  \label{al:1}
  \SetKwData{Left}{left}\SetKwData{This}{this}\SetKwData{Up}{up}
  \SetKwFunction{Union}{Union}\SetKwFunction{FindCompress}{FindCompress}
  \SetKwInOut{Input}{Input}\SetKwInOut{Output}{Output}
  \small
  \Input{The feature of $i$-th sample $x_i$ with its label $y_i$, last fully-connected layer parameters $W$, cosine similarity $\cos {\theta _j}$ of two vectors, embedding network parameters $\Theta $, and margin $m$}
  iteration number $k \leftarrow 0$, parameter $m \leftarrow 0.5$, $\lambda \leftarrow \pi$;\\
  \While {not converged} {
  for all $j \ne {y_i}$,
  
  \eIf{$\max \{ \cos ({\theta _j} + m)\}  > \cos {\theta _{{y_i}}}$}{${y_i} = j$;}{${y_i} = {y_i}$;}
  
  \eIf{$\cos ({\theta _{{y_i}}} + m) > \max \{ \cos {\theta _j}\}$}{$f = 0$;}{$f = \max \{ \cos {\theta _j}\} - \cos ({\theta _{{y_i}}} + m)$;}
  
  Compute the loss $L$ by Eq. \ref{eq:8};
  
  Compute the gradients of $x_i$ and $W_j$ by Eq. \ref{eq:9},\ref{eq:10},\ref{eq:11},\ref{eq:12};
  
  Update the parameters $W$ and $\Theta$;
  
  $k \leftarrow k + 1$;
  }
  \Output{$W$, $\Theta$}
\end{algorithm}
% \begin{figure}[h]
%     \centering
%     \includegraphics[scale=0.4]{loss_curve_v2.pdf}
%     \caption{Convergence process of BoundaryFace.}
%     \label{fig:my_label4}
% \end{figure}
\begin{figure}[h]
    \centering
    \includegraphics[scale=0.3]{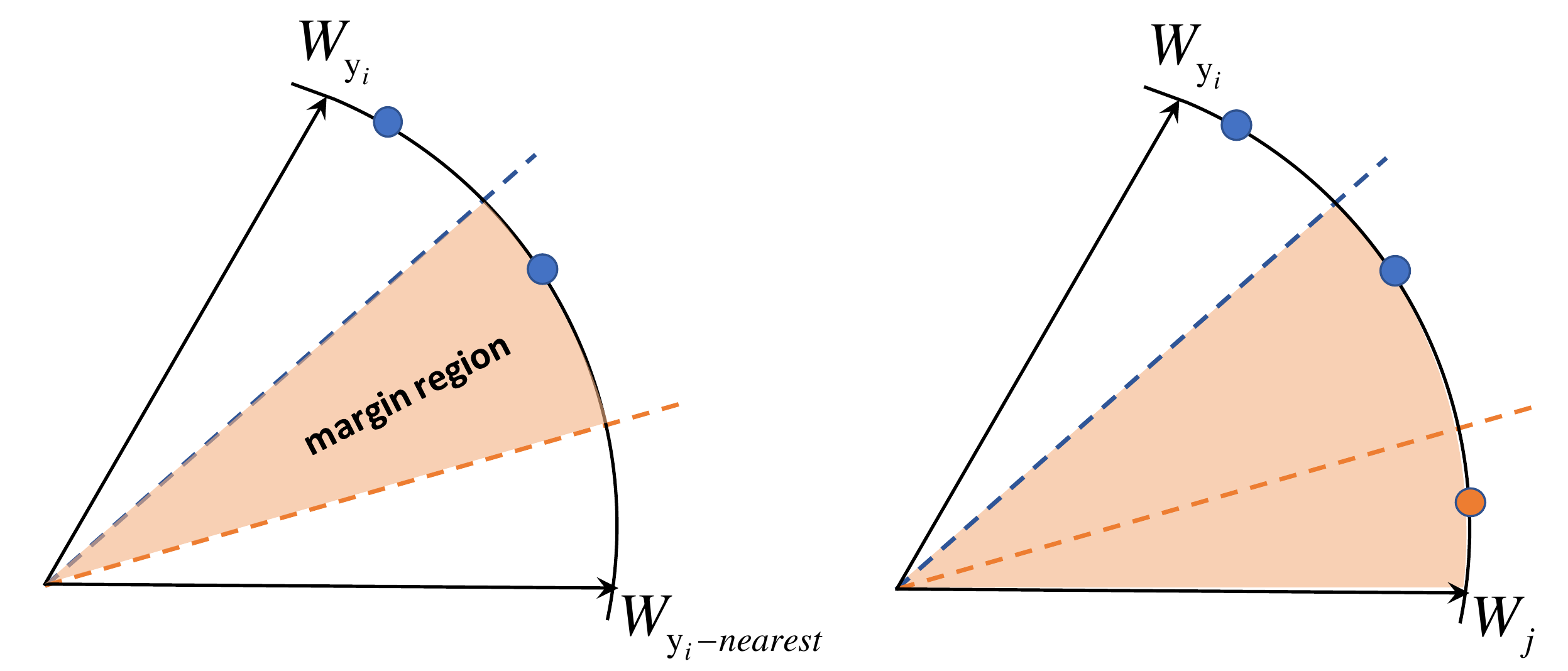}
    \caption{The difference between BoundaryFace and SOTA methods in emphasizing the region where the hard samples are located.
    %{\bf Left:} The light orange area represents the margin region where the hard samples highlighted by our method are located. {\bf Right:} The light orange area represents the region where the mis-classified samples reinforced by SOTA methods are situated.
    }
    \label{fig:my_label5}
\end{figure}
%------------------------------------------------------------------------
\subsection{Discussions with SOTA Loss Functions}
\noindent{\bf Comparison with baseline softmax.} The baseline softmax (\textit{i.e.}, ArcFace, CosFace) introduces a margin from the perspective of positive cosine similarity, and they treat all samples equally. Our approach mining hard sample by introducing a regularization term to make the network pay extra attention to the hard samples.

~\\
{\bf Comparison with MV-Arc-Softmax and CurricularFace.} MV-Arc-Softmax and CurricularFace assign the same semantics to hard samples, only differing in the following handling stage. As shown in Fig. \ref{fig:my_label5} (right), both treat mis-classified samples as hard samples and focus on the relationship between the ground truth class of the sample and all negative classes. Instead, in Fig. \ref{fig:my_label5} (left), our approach regards these samples located in the margin region as hard samples and focuses only on the relationship between the ground truth class of the sample and the nearest negative class. Moreover, the labels of both blue and orange samples are $y_i$, but the orange sample ground truth class is $j$-th class. Obviously, if there is closed-set noise in the dataset, SOTA methods not only emphasize hard samples but also reinforce closed-set noise.

%------------------------------------------------------------------------
% 实验数据1
\section{Experiments}
\subsection{Implementation Details}
\noindent{\bf Datasets.} CASIA-WebFace \cite{yi2014learning} which contains about 0.5M of 10K individuals, is the training set that is widely used for face recognition, and since it has been cleaned very well, we take it as a clean dataset.  In order to simulate the situation where the dataset contains much of noise, based on the CASIA-WebFace, we artificially synthesize noisy datasets which contain a different ratio of noise. In detail, for closed-set noise, we randomly flip the sample labels of CASIA-WebFace; for open-set noise, we choose MegaFace \cite{kemelmacher2016megaface} as our open-set noise source and randomly replace the samples of CASIA-WebFace. Finally, we use the clean CASIA-WebFace and noisy synthetic datasets as our training set, respectively. We extensively test our method on several popular benchmarks, including LFW \cite{huang2008labeled}, AgeDB \cite{moschoglou2017agedb}, CFP-FP \cite{sengupta2016frontal}, CALFW \cite{zheng2017cross}, CPLFW \cite{zheng2018cross}, SLLFW \cite{deng2017fine}, RFW \cite{wang2018racial}.  RFW consists of four subsets: Asian, Caucasian, Indian, and African. Note that in the tables that follow, CA denotes CALFW, CP denotes CPLFW, and Cau denotes Caucasian.

~\\
{\bf Training Setting.} We follow \cite{deng2019arcface} to crop the 112×112 faces with five landmarks \cite{zhang2016joint} \cite{tai2019towards}. For a fair comparison, all methods should be the same to test different loss functions. To achieve a good balance between computation and accuracy, we use the ResNet50 \cite{he2016deep} as the backbone. The output of backbone gets a 512-dimension feature. Our framework is implemented in Pytorch \cite{paszke2017automatic}. We train modules on 1 NVIDIA TitanX GPU with batch size of 64. The models are trained with SGD algorithm, with momentum 0.9 and weight decay 5$e$ -4. The learning rate starts from 0.1 and is divided by 10 at 6, 12, 19 epochs. The training process is finished at 30 epochs. We set scale $s$  = 32 and margin $m$ = 0.3 or $m$ = 0.5. Moreover, to make the network with sufficient discrimination ability, we first pre-train the network for 7 epochs using margin-based loss. The margin-based loss can also be seen as a degenerate version of our BoundaryFace.
%------------------------------------------------------------------------
\begin{table}[t]
    \centering
     \caption{ Verification performance (\%) of our BoundaryFace with different hyper-parameter $\lambda$.}
    \begin{tabular}{l | c  r}
       \hline
       Method & SLLFW & CFP-FP  \\
       \hline \hline
       $\lambda = 2$  & 98.05 & 94.8 \\
       $\lambda = 2.5$  & 97.9 & 94.71 \\
       $\lambda = \pi$  & {\bf 98.12} & {\bf 95.03} \\
       $\lambda = 3.5$  & 97.8 & 94.9 \\
       \hline
    \end{tabular}
    % caption
    \label{tab:lambda effect}
\end{table}
% m实验
\begin{table}[t]
    \centering
    \caption{ Verification performance (\%) of ArcFace with different hyper-parameter $m$ on datasets which contain different noise mixing ratios (\%).}
    \begin{tabular}{l | c  c  c  r}
       \hline
       $m$ & closed-set ratio & open-set ratio & LFW & AgeDB \\
       \hline \hline
       $0.3$  & 10\% & 30\%  & {\bf 99.07} & {\bf 91.82} \\
       $0.5$  & 10\% & 30\% & 98.22 & 89.02 \\
       \hline
       $0.3$  & 30\% & 10\% &  98.42 & 88.5 \\
       $0.5$  & 30\% & 10\% & {\bf 98.73} & {\bf 89.93} \\
       \hline
    \end{tabular}
    \label{tab:m effect}
\end{table}
\subsection{Hyper-parameters}
% lambda实验
\noindent{\bf Parameter $\lambda$.} Since the hyper-parameter $\lambda$ plays an essential role in the proposed BoundaryFace, we mainly explore its possible best value in this section. In Tab. \ref{tab:lambda effect}, we list the performance of our proposed BoundaryFace with $\lambda$ varies in the range [2, 3.5]. We can see that our BoundaryFace is insensitive to the hyper-parameter $\lambda$. And, according to this study, we empirically set $\lambda$ = $\pi$.

~\\
{\bf Parameter $m$.} Margin $m$ is essential in both margin-based softmax loss and mining-based softmax loss. For clean datasets, we follow \cite{deng2019arcface} to set margin $m$ = 0.5. In Tab. \ref{tab:m effect}, we list the performance of different $m$ for ArcFace on datasets with different noise ratios. It can be concluded that if most of the noise in the training set are open-set noise, we set $m$ = 0.3; otherwise, we set $m$ = 0.5.
%------------------------------------------------------------------------
\subsection{Comparisons with SOTA Methods}
\begin{figure}[h]
    \centering
    \includegraphics[scale=0.32]{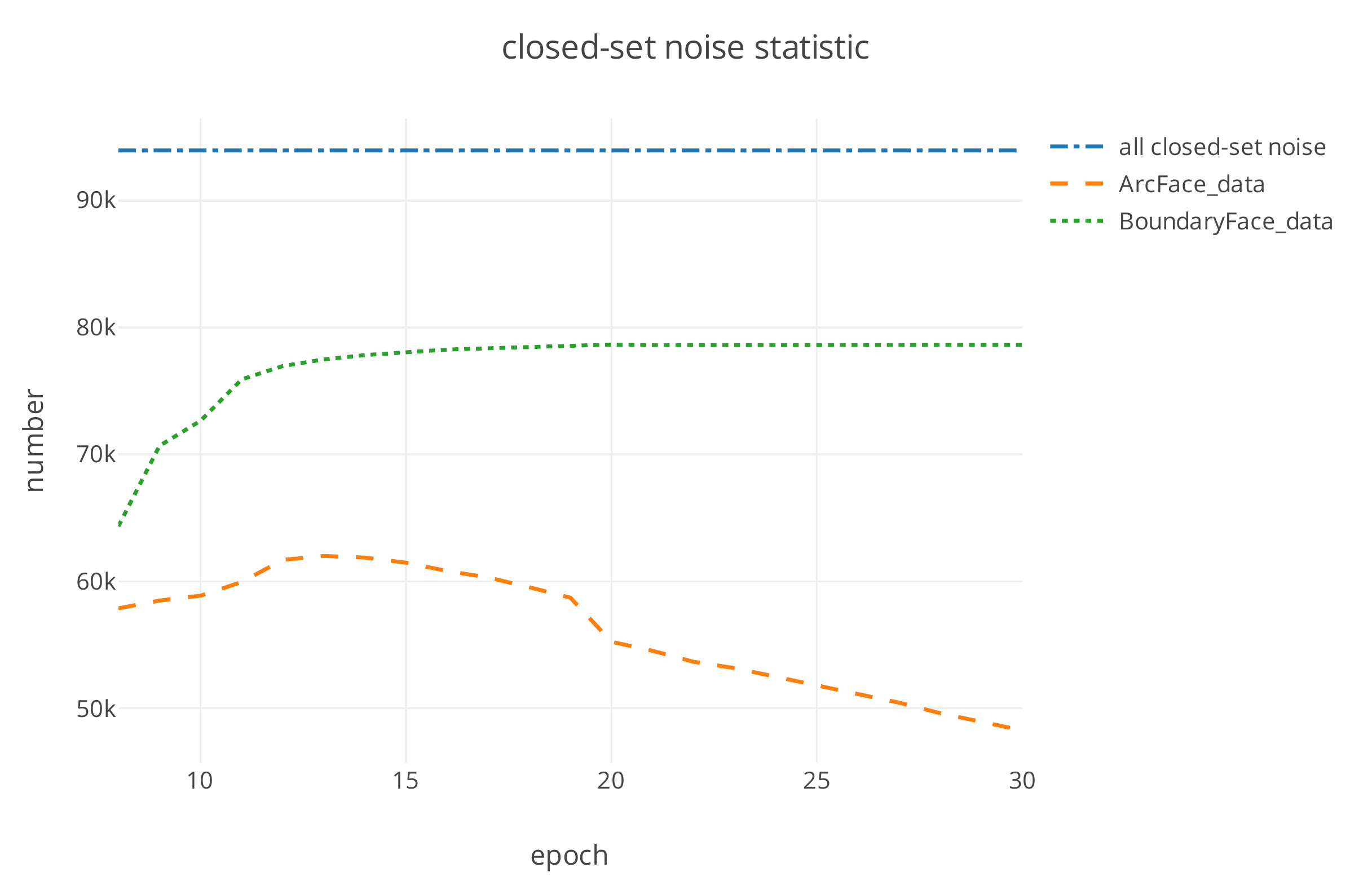}
    \caption{Comparison of closed-set noise detected by BoundaryFace and ArcFace, respectively. The dash dot line indicates the total number of closed-set noise in the training set.}
    \label{fig:noise_number}
\end{figure}

\noindent{\bf Results on a dataset that is clean or contains only closed-set noise.} In this section, we first train our BoundaryFace on the clean dataset as well as datasets containing only closed-set noise. We use BoundaryF1 (Eq. \ref{eq:7}) as a reference to illustrate the effects of hard sample mining. Tab. \ref{tab:expriment1} provides the quantitative results. It can be seen that our BoundaryFace outperforms the baseline and achieves comparable results when compared to the SOTA competitors on the clean dataset; our method demonstrates excellent superiority over baseline and SOTA methods on closed-set noise datasets. Furthermore, we can easily draw the following conclusions: 1) As the closed-set noise ratio increases, the performance of every compared baseline method drops quickly; this phenomenon did not occur when using our method. 2) Mining-based softmax has the opposite effect when encountering closed-set noisy data, and the better the method performs on the clean dataset, the worse the results tend to be. In addition, in Fig. \ref{fig:noise_number}, given 20\% closed-set noise, we present the detection of closed-set noise by BoundaryFace during training and compare it with ArcFace. After closed-set noise is detected, our BoundaryFace dynamically corrects its labels. Correct labels result in a shift in the direction of the closed-set noise being optimized from wrong to right, and also lead to more accurate class centers. Furthermore, more accurate class centers in turn allow our method to detect more closed-set noise at each iteration, eventually reaching saturation.
% 实验1
\begin{table*}[pt]
    \centering
    \caption{Verification performance (\%) of different loss functions when the training set contains different ratios of closed-set noise. Ratio 0\% means that the training set is the original CASIA-WebFace (\textit{i.e.}, clean dataset).}
    \scriptsize
    \begin{tabular}{c| c | c c c c c c c c c c}
        \hline
        ratio & Method & LFW & AgeDB & CFP & CA & CP & SLLFW & Asian
& Cau & Indian & African \\
        \hline \hline
        \multirow{5}{*}{0\%}  & ArcFace & 99.38 & 94.05 & 94.61 & 93.43 & 89.45 & 97.78 & 86.5 & 93.38 & 89.9 & 86.72 \\
          & MV-Arc-Softmax & 99.4 & 94.17 & 94.96 & 93.38 & 89.48 & 97.88 & 86.23 & 93.27 & 90.12 & 87.03 \\
         & CurricularFace & 99.42 & 94.37 & 94.94 & {\bf 93.52} &  89.7 & 98.08 & 86.43 & {\bf 94.05} & {\bf 90.55} & {\bf 88.07} \\
          & BoundaryF1 & 99.41 & 94.05 & 95.01 & 93.27 & {\bf 89.8} & 97.75 & 85.72 & 92.98 & 89.98 & 86.43 \\
          & {\bf BoundaryFace} & {\bf 99.42} & {\bf 94.4} & {\bf 95.03} & 93.28 & 89.4 & {\bf 98.12} & {\bf 86.5} & 93.75 & 90.5	& 87.3 \\
         \hline
         \multirow{5}{*}{10\%}  & ArcFace & 99.33 & 93.81 & 94.34 & 93.11 & 89.1 & 97.67 & 85.87 & 92.98 & 90.15 & 86.52 \\
         & MV-Arc-Softmax & {\bf 99.43} & 93.9 & 94.27 & 93.15 & {\bf 89.47} & 97.82 & 85.7 & 93.17 & {\bf 90.45} & 87.28 \\
         & CurricularFace & 99.33 & 93.92 & 93.97 & 93.12 & 88.78 & 97.52 & 85.43 & 92.98 & 89.53 & 86.53 \\
         & BoundaryF1 & 99.4 & 94.02 & 94.3 & 93.18 & 89.32 & 97.85 & 86.5 & 93.32 & 90.33 & 86.95 \\
         & {\bf BoundaryFace} & 99.35 & {\bf 94.28} & {\bf 94.79} & {\bf 93.5} &  89.43 & {\bf 98.15} & {\bf 86.55} & {\bf 93.53} &  90.33 & {\bf 87.37} \\
         \hline
         \multirow{5}{*}{20\%} & ArcFace & 99.3 & 93 & 93.49 & 92.78 & 88.12 & 97.57 & 84.85 & 91.92 & 88.82 & 85.08 \\
          & MV-Arc-Softmax & 99.12 & 93.12 & 93.26 & 93.12 & 88.3 & 97.37 & 85.15 & 92.18 & 89.08 & 85.32 \\
          & CurricularFace & 99.13 & 91.88 & 92.56 & 92.28 & 87.17 & 96.62 & 84.13 & 91.13 & 87.7 & 83.6 \\
          & BoundaryF1 & 99.32 & 94.02 & {\bf 94.5} & 93.18 & {\bf 89.03} & 97.63 & 86 & {\bf 93.28} & 89.88 & 86.48 \\
          & {\bf BoundaryFace} & {\bf 99.38} & {\bf 94.22} & 93.89 & {\bf 93.4} &  88.45 & {\bf 97.9} & {\bf 86.23} & 93.22 & {\bf 90} & {\bf 87.27} \\
         \hline
    \end{tabular}
    \label{tab:expriment1}
\end{table*}

~\\
% 实验数据2
{\bf Results on noisy synthetic datasets.} The real training set contains not only closed-set noise but also open-set noise. As described in this section, we train our BoundaryFace on noisy datasets with different mixing ratios and compare it with the SOTA competitors. In particular, we set the margin $m$ = 0.5 for the training set containing closed-set noise ratio of 30\% and open-set noise ratio of 10\%, and we set the margin $m$ = 0.3 for the other two mixing ratios. As reported in Tab. \ref{tab:expriment2}, our method outperforms the SOTA methods on all synthetic datasets. Even on the dataset containing 30\% open-set noise, our method still performs better than baseline and SOTA competitors.
% Further, we can also conclude that neural networks are more sensitive to closed-set noise than open-set noise. The noise problem can be alleviated for a large-scale face dataset if we effectively solve the closed-set noise problem.
% 实验数据2
\begin{table*}[pt]
    \centering
    \caption{Verification performance (\%) of different loss functions when the training set contains different mixing noise ratios. C denotes closed-set noise ratio(\%). O denotes open-set noise ratio(\%).}
    \scriptsize
    \begin{tabular}{c | c | c | c c c c c c c c c c}
        \hline
        C & O & Method & LFW & AgeDB & CFP & CA & CP & SLLFW & Asian
& Cau & Indian & African \\
        \hline \hline
        \multirow{5}{*}{20}  & \multirow{5}{*}{20} &  ArcFace & 98.87 & 89.55 & 89.56 & 90.43 & 84.4 & 94.87 & 80.97 & 88.37 & 85.83 & 79.72 \\
         & & MV-Arc-Softmax & 98.93 & 89.95 & 89.61 & 91.67 & 84.9 & 95.67 & 82.25 & 88.98 & 86.22 & 80.6 \\
        & & CurricularFace & 98.33 & 88.07 & 88.29 & 90.18 & 83.65 & 93.87 & 80.3 & 87.78 & 84.43 & 77.35 \\
        & & BoundaryF1 & 99.12 & 92.5 & 89.66 & 92.27 & 84.43 & 96.75 & 84.42 & 91.4 & 88.33 & {\bf 84.55}  \\ 
        & & {\bf BoundaryFace} & {\bf 99.2} & {\bf 92.68} & {\bf 93.28} & {\bf 92.32} & {\bf 87.2} & {\bf 96.97} & {\bf 84.78} & {\bf 91.88} & {\bf 88.8} & 84.48 \\
        \hline
        \multirow{5}{*}{10}  & \multirow{5}{*}{30} &  ArcFace & 99.07 & 91.82 & 91.34 & 91.7 & 85.6 & 96.45 & 83.35 & 90 & 87.33 & 81.83 \\
         & & MV-Arc-Softmax & 98.92 & 91.23 & 91.27 & 91.9 & 85.63 & 96.17 & 83.98 & 90.15 & 87.57 & 82.33 \\
         & & CurricularFace & 98.88 & 91.58 & 91.71 & 91.83 & 85.97 & 96.17 & 82.98 & 89.97 & 86.97 & 82.13 \\ 
         & & BoundaryF1 & 99 & 92.23 & 92.07 & {\bf 92.05} & {\bf 86.62} & 96.42 & 83.97 & 90.47 & 87.83 & 82.98 \\
         & & {\bf BoundaryFace} & {\bf 99.17} & {\bf 92.32} & {\bf 92.4} & 91.95 & 86.22 & {\bf 96.55} & {\bf 84.15} & {\bf 91.05} & {\bf 88.17} & {\bf 83.28}\\
        \hline
        \multirow{5}{*}{30}  & \multirow{5}{*}{10} & ArcFace & 98.73 & 89.93 & 89.21 & 91.02 & 82.93 & 95.15 & 81.25 & 88.33 & 85.87 & 80.03 \\
         & & MV-Arc-Softmax & 98.78 & 89.73 & 88.54 & 91.22 & 82.57 & 95.22 & 81.52 & 88.47 & 85.83 & 80.12 \\
         & & CurricularFace &  98.18 & 87.65 & 88.1 & 90.12 & 82.82 & 93.13 & 79.7 & 86.65 & 84.1 & 77.22 \\ 
         & & BoundaryF1 & 99.1 & 92.3 & {\bf 90.34} & 92.28 & {\bf 85.28} & 96.7 & 83.52 & 90.9 & 87.53 & 83.12 \\
         & & {\bf BoundaryFace} & {\bf 99.1} & {\bf 93.38} & 88.24 & {\bf 92.5} & 82.22 & {\bf 96.88} & {\bf 83.77} & {\bf 91.18} & {\bf 88.18} & {\bf 83.67} \\
        \hline
    \end{tabular}
    \label{tab:expriment2}
\end{table*}
%------------------------------------------------------------------------
\section{Conclusions}
In this paper, we propose a novel mining framework (\textit{i.e.}, BoundaryFace) with tolerating closed-set noise for face recognition. BoundaryFace largely alleviates the poor performance of mining-based softmax on datasets with severe noise problem. BoundaryFace is easy to implement and converges robustly. Moreover, we investigate the effects of noise samples that might be optimized as hard samples. Extensive experiments on popular benchmarks have demonstrated the generalization and effectiveness of our method when compared to the SOTA.
%------------------------------------------------------------------------
\section*{Acknowledgement}
This work was supported in part by the National Natural Science Foundation of China (61876158), Fundamental Research Funds for the Central Universities (2682021ZTPY030).

\clearpage
% \bibliography{egbib}

\end{document}